\RequirePackage{fix-cm}
\documentclass[smallextended]{svjour3}       
\smartqed  

\usepackage{graphicx}
\usepackage{cite}
\usepackage{wrapfig}
\usepackage{amsfonts}
\usepackage{mathtools}
\usepackage{amsmath}
\usepackage{algpseudocode}
\usepackage{algorithm}
\usepackage{multirow}

\DeclareMathOperator*{\argmin}{arg\,min}

\iftrue
  \paperwidth=\dimexpr
    1in + \oddsidemargin
    + \textwidth
    + 1in + \oddsidemargin
  \relax
  \paperheight=\dimexpr
    1in + \topmargin
    + \headheight + \headsep
    + \textheight
    + 1in + \topmargin
  \relax
  \usepackage[pass]{geometry}\relax
\fi

\begin{document}

\title{PIAT: Physics Informed Adversarial Training for Solving Partial Differential Equations
}

\titlerunning{PIAT: Physics Informed Adversarial Training for Solving PDEs.}        
\authorrunning{S. Shekarpaz, M. Azizmalayeri, MH. Rohban}

\author{Simin Shekarpaz* \and \\ 
Mohammad Azizmalayeri* \and \\ 
Mohammad Hossein Rohban
}

\institute{\textbf{* These authors contributed equally to this work.} \\ \\
S. Shekarpaz  \quad
              \email{siminshekarpaz@gmail.com}  \\ 
           M. Azizmalayeri  \quad
              \email{m.azizmalayeri@sharif.edu} \\
        MH. Rohban \quad
                \email{rohban@sharif.edu}\\ \at
                Department of Computer Engineering, Sharif University of Technology, Tehran, Iran
}

\date{Received: date / Accepted: date}

\maketitle

\begin{abstract}
In this paper, we propose the physics informed adversarial training (PIAT) of neural networks for solving nonlinear differential equations (NDE). It is well-known that the standard training of neural networks results in non-smooth functions. Adversarial training (AT) is an established defense mechanism against adversarial attacks, which could also help in making the solution smooth. AT include augmenting the training mini-batch with a perturbation that makes the network output mismatch the desired output adversarially. Unlike formal AT, which relies only on the training data, here we encode the governing physical laws in the form of nonlinear differential equations  using automatic differentiation in the adversarial network architecture.
We compare PIAT with PINN to indicate the effectiveness of our method in solving NDEs for up to 10 dimensions. Moreover, we propose weight decay and Gaussian smoothing to demonstrate the PIAT advantages. The code repository is available at https://github.com/rohban-lab/PIAT.
\end{abstract}

\section{Introduction}
The efficiency of deep learning has recently proven to be significant in various scientific fields \cite{1}, such as computer vision, natural language processing, robotics, and physics simulation. One of the effective applications of deep neural networks is to approximate the solution of a physical system \cite{2, 3, 4, 5, 6}.

The idea of using neural networks to solve ordinary and partial differential equations was first studied by researchers in \cite{7, 8, 9, 10, 11}, where the approximate solution is parameterized by a fully-connected network that allows for a fully differentiable and closed analytic form. Applications of these techniques in scientific computation and numerical approximation of physical systems has increased due to the widespread interest in neural networks.

In \cite{12}, Raissi introduced a physics-informed neural network (PINN) in which the underlying PDE and boundary conditions are enforced through minimization of the loss functions, and the solution derivatives are obtained using the automatic differentiation. The points on the boundary conditions are also considered as the training set.

Efficiency of different types of PINNs has been shown \cite{14, 15, 16} to solve different classes of PDEs, such as integro differential equations \cite{17},  fractional equations \cite{17},  surfaces PDEs \cite{18}, and stochastic differential equations \cite{19}. In \cite{20}, a conservative physics-informed neural network (cPINN) has been developed to satisfy various conservation laws while solving the PDEs, where the problem is solved in multiple subdomains and ensure continuity of the flux across subdomains. This method can also be used to solve the weak form (variational) of a PDE. Since the weak form contains natural boundary conditions, the neural network solution should only meet the necessary boundary conditions. In \cite{Karumuri2020SimulatorfreeSO}, the authors considered the variational form for stochastic PDEs and applied the idea of PINN to obtain the PDE solution. In addition, a variational formulation of PINNs (hp-VPINN) was proposed to deal with PDEs with non-smooth solutions \cite{21}, which is based on the Galerkin method. A similar version was also studied in XPINN \cite{22}; a  domain decomposition approach in the PINN framework that conforms to the conservation laws. The authors in \cite{23} proposed a Bayesian approach to physics-informed neural networks to solve the forward and inverse problems. A parallel framework for the domain decomposition of cPINNs and XPINNs was also developed in \cite{24}. The Python library DeepXDE \cite{lu2021deepxde}, is an implementation of PINNs,  which makes the user's codes compact, and follow closely the mathematical formulation. 

Some other recent works have focused on the development of neural network architecture and training \cite{25, 26, 27, 28, 29} that can improve the performance of PINNs in various fields. Furthermore, in \cite{30, 31, 32}, the error estimates and convergence of PINN methods were discussed.

In this paper, we explore effectiveness of the neural network smoothing methods in the performance of PINNs. Specifically, we first propose weight decay \cite{Krogh91} and Gaussian smoothing as possible methods that may help the neural network model smoothness, and hence generalization. Weight decay adds a penalty term to the loss function that promotes smaller weight norms in the model, and reduces potential overfitting of the model. Gaussian smoothing adds a small Gaussian random perturbation to each training sample, and performs similar to an augmentation on the input samples. Next, we propose a more generalized method for solving nonlinear problems based on adversarial training called PIAT. Adversarial training has been proved to be effective against the adversarial examples, where small adversarial perturbations are added to the original input such that model produces incorrect outputs \cite{SzegedyZSBEGF13, NguyenYC15, Madry2018}. By using this method, the fully connected neural network is trained to solve physical problems in various dimensions, which leads to better robustness, and generalization of the neural network on the test data, while PINN might not be effective enough. This model is implemented for three PDE systems, Kuramoto-Sivashinsky equation, Sawada–Kotera equation and high dimensional Allen-Cauhn equation. In all cases, the total number of training data is relatively small and the train and test data are generated using Latin Hypercube Sampling (LHS) strategy. The loss functions are minimized using the Adam optimization algorithm.

In the proposed PIAT method, a single neural network is used for the whole domain, which automatically enforces continuity of the solution and its derivatives over time. By using this method, fewer iterations and collocation points are used to achieve convergence in comparison with the standard PINN. The numerical results show that the PIAT method works for high order and nonlinear PDEs and its performance is significantly better than the PINN or other methods such as the Gaussian smoothing or weight decay.

The remainder of this paper is organized as follows. In section 2, we review the PINN formulation to solve differential equations.
Section 3 introduces weight decay and Gaussian smoothing as baselines for the generalization purpose.
The proposed adversarial training method (PIAT) is  introduced in section 4 and their properties are also presented.
In section 5, PIAT is applied to 3 different problems and compared with other methods to demonstrate the efficiency of the proposed technique.

\section{Physics-informed neural networks (PINNs) for solving PDEs\label{sec:PINN}}



We consider the general form of an $m$-th order initial value problem (IVP), which is as follows:\\
\begin{equation}\label{eq1}
\begin{split}
&u_t(\textbf{x}, t)+N[u(\textbf{x}, t)] = 0,\quad \textbf{x} \in \Omega, \ t \in [0, T],\\
&u(\textbf{x}, 0)=h(\textbf{x}), \quad \textbf{x} \in \Omega,\\
&u(\textbf{x}, t)=g(\textbf{x}, t), \quad \textbf{x} \in \partial\Omega, \ t \in [0, T],
\end{split}
\end{equation}
where $N$ represents a nonlinear differential operator, $\lambda$ is a parameter, $\Omega \subseteq \mathbb{R}^d$ and $u$ is the unknown solution with known initial and boundary conditions. Moreover, $u_t({\mathbf x}, t) = \partial u / \partial t$.  

By using PINN framework, a fully connected neural network, which is composed of multiple hidden layers, is used to approximate the solution $u(\textbf{x}, t)$ of the given nonlinear problem. The inputs of the network are $(\textbf{x}, t)$ and the outputs are considered as $\hat{u}(\textbf{x}, t)$. 
Then, each hidden neuron of the neural network in the $l$-th layer can be expressed as
\begin{equation}\label{eq10}
y^{(l)}_j= \sigma\left(\sum_i w^{(l)}_{i, j} {x}^{(l)}_i+b^{(l)}_j\right),
\end{equation}
where $x^{(l)}_i$'s are the inputs in the $l$-th layer, and $y^{(l)}_j$'s are the outputs of each hidden neuron in the $l$-th layer. The output of the last layer, $y^{(L)}_1$ is used to approximate the solution. In this formulation $w^{(l)}_{i, j}$ and $b^{(l)}_j$ are the trainable weights and biases in the $l$-th layer, and $\sigma(.)$ is the continuous activation function.

The parameters of the neural network are randomly initialized and iteratively updated by minimizing the loss function that consists of the residuals of three error terms including the PDE, initial condition, and boundary conditions:
\begin{equation} \label{loss_func}
\begin{split}
\mathrm{minimize}_{{\mathbf w}, {\mathbf b}} \quad \quad  &\frac{1}{N_r}  \sum_{i=1}^{N_r} \vert u_t( \textbf{x}_i, t_i)+N[u(\textbf{x}_i, t_i)] \vert^2\\
&+ \frac{1}{N_b}  \sum_{i=1}^{N_b} \vert u( \textbf{x}_i, t_i) - g({\mathbf x}_i, t_i) \vert^2+\frac{1}{N_0}  \sum_{i=1}^{N_0} \vert u( \textbf{x}_i, 0) - h({\mathbf x}_i) \vert^2,
\end{split}
\end{equation}
where $u(\textbf{x}, t)$ is the estimated solutions and $N_f$, $N_b$, and $N_0$ denote the number of collocation points, boundary points, and initial points, respectively, which can be chosen arbitrary. By solving the minimization problem, the optimal value of the weights ${\mathbf w}^\star$ and biases ${\mathbf b}^\star$ are computed and then the approximate solutions $\hat{u}(\textbf{x}, t) = u({\mathbf x}, t; {\mathbf w}^\star, {\mathbf b}^\star)$ is obtained.

In the PINN formulation, the solution’s derivatives are computed using the automatic differentiation, where combining the derivatives of the constituent operations by the chain rule gives the derivative of the overall composition.
The capabilities of automatic differentiation are well-employed in most deep learning frameworks such as TensorFlow and PyTorch, and it allows us to avoid time consuming derivations or numerical discretization while computing derivatives of all orders in space-time.

\section{Gaussian smoothing and weight decay}
To improve the generalization of neural network in solving the physics informed equations, we propose to implicitly make the solution robust to small random perturbations. Such robustness is known to cause model generalization, which is also desired in the case of PINNs. To investigate the effectiveness of this idea, we add a random bounded noise to the input as an augmentation in the training, and also apply weight decay in order to reach a stable model that generalizes well on the test data.

Augmentation is a well-known technique in training of the machine learning models, as it prevents the model from overfitting to the training data. For instance, rotation and crop are broadly used in the context of image classification models \cite{ShortenK19, Luke18}. In this problem, we can not use geometrical transforms due to the data types, but we can add random perturbations to the samples. Therefore, we apply Gaussian smoothing in the training as an augmentation.
Perturbing input samples with Gaussian noise can be simply done as:
\begin{equation}\label{}
x_{Gaussian} = x + \mathcal{N}(0,\,\sigma^{2}),
\end{equation}
where Gauussian noise with mean equal to $0$ and variance equal to $\sigma^{2}$ is added to the input $x$.

Furthermore, we use the weight decay \cite{Krogh91} as another effective method in preventing overfitting and improving generalization. Weight decay is a regularization term that adds a penalty term to the loss function to keep the model weights small as follows:
\begin{equation}
    loss = \mathcal{L}(f_\theta, y) + \lambda\ \|\theta\|^2,
\end{equation}
where $\lambda$ is the hyper-parameter that controls the impact of the weight decay. It is clear that a large $\lambda$ causes higher regularization effect, which leads to more smooth solutions. Likewise, higher $\sigma$ in the Gaussian noise augmentation have a similar effect.

\section{Physics-Informed Adversarial Training of Neural Networks (PIAT)}\label{sec_PIAT}

Using a training set $\chi$ and a neural network $f_\theta(.)$, with weights $\theta$, the standard training can be done as:
\begin{equation}\label{}
\theta^* = \argmin_\theta \ \mathbb{E}_{(({\mathbf x}, t),y)\in\chi} \ \ell(({\mathbf x}, t),y; f_\theta),
\end{equation}
to minimize the loss $\ell(.)$ of $f_\theta(.)$, the estimated PDE solution, on training samples. Note that the loss function $\ell(.)$ is defined based on the Eq. \ref{loss_func}:
\begin{equation}
\ell(({\mathbf x}, t), y; f_\theta) := \begin{cases} (\partial f_\theta / \partial t + N[f_\theta({\mathbf x}, t)] - y_1)^2; ~~~   ({\mathbf x}, t) \in \mathcal{C} \\ (f_\theta({\mathbf x}, t) - y_2)^2; ~~~ ({\mathbf x}, t) \in \mathcal{B} \\ (f_\theta({\mathbf x}, 0) - y_3)^2; ~~~ ({\mathbf x}, 0) \in \mathcal{I},
    \end{cases}
\end{equation}
where $y_1 = 0$, $y_2 = g({\mathbf x}, t)$, and $y_3 = h({\mathbf x})$, for the training sample $({\mathbf x}, t)$, and $\mathcal{C}$, $\mathcal{B}$, and $\mathcal{I}$ denote collocation, boundary, and initial points.
While standard training performs well in many tasks, it leads to models that are fragile against adversarial examples. These examples are defined as the addition of an $\ell_p$-bounded perturbation to the original input such that model generates outputs with large errors \cite{SzegedyZSBEGF13, NguyenYC15}. This issue is due to the fact that the model does not learn robust features for solving the problem \cite{IlyasSTETM19, d2020underspecification}. The fact that the training data does not  sufficiently cover the input space suggests that the physics-informed neural networks also suffer from the same phenomenon. The reason for using weight-decay and Gaussian smoothing was to mitigate this issue. However, they might not be as effective as they should be.

Another solution for this problem is the adversarial training that is widely explored on the image classification task \cite{Madry2018, ZhangYJXGJ19, ZhangXH0CSK20, az2021}. Adversarial training has two main parts. In the first part, a small perturbation $\delta$ is optimized and added to the training sample ${\mathbf x}$ such that the network loss is maximized. Next, the neural network weights are optimized based on a single stochastic gradient update of the loss function defined on the perturbed sample $({\mathbf x}, t) + \delta$. As a result, the neural network can learn more robust features and perform well even against adversarially perturbed samples. Adversarial training is summarized as:
\begin{equation}\label{AT}
\theta^* = \argmin_\theta \ \mathbb{E}_{(({\mathbf x}, t),y)\in\chi} \ \left(\max_{(\delta_x, \delta_t) \in [-\epsilon, \epsilon]^{d+1}} \ell(({\mathbf  x}+\delta_x, t + \delta_t), y; f_\theta)\right),
\end{equation}
where $\epsilon$ is the $\ell_\infty$ bound for the $\delta$. We assume that $\delta$ constitutes two parts $\delta_x$ and $\delta_t$, which perturb ${\mathbf x}$ and $t$, respectively. The minimization part of Eq. \ref{AT} is done with the stochastic Gradient Descent (SGD). For the maximization part, different algorithms can be applied but we use Projected Gradient Descent (PGD) \cite{Madry2018}. PGD attack is an iterative algorithm that computes gradients of the loss with respect to the input, and uses their sign to generate perturbations that maximize the model loss as the following:
\begin{equation}\label{}
\delta_{i+1} = \delta_{i} + \alpha \ sign(\nabla_{(\mathbf x, t)} \ell(({\mathbf x}, t), y; f_\theta)),
\end{equation}
where $\alpha$ is the size that we move along the gradient sign direction in each step. 

\begin{figure}
    \centering
    \includegraphics[width=\textwidth]{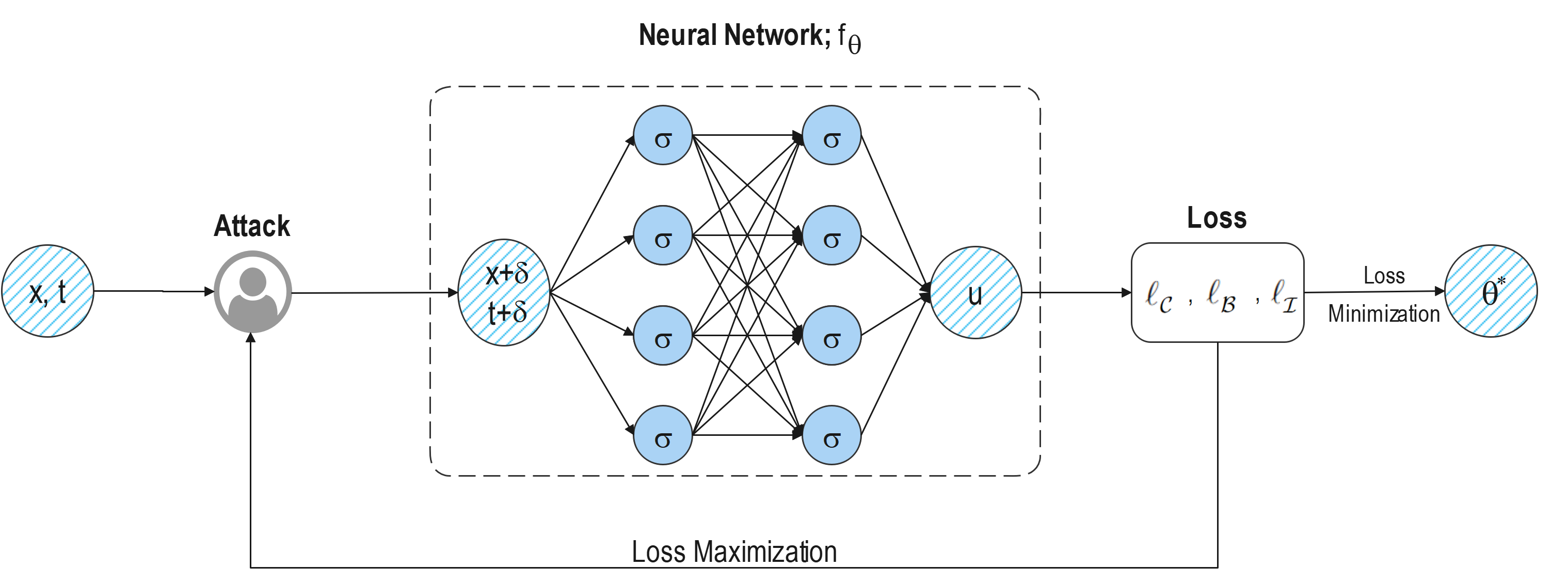}
    \caption{PIAT method schematic. The neural network is optimized based on the adversarially perturbed inputs. These inputs are crafted via adding a bounded perturbation to the original ones in order to maximize the neural network loss.}
    \label{fig:method}
\end{figure}

The pure adversarial training itself does not work well in PIAT due to the fact that perturbing training samples in the physics-informed neural networks changes the correct label, $y$, for the sample much more than perturbing images in the vision tasks. Therefore, in PIAT, we propose to optimize the model on ($({\mathbf x}, t)+\delta$, $y'$), where $y'$ is the ground truth label for the ${\mathbf x}+\delta$, where 
\begin{equation}
y' = (0, g({\mathbf x} + \delta_x, t + \delta_t), h({\mathbf x} + \delta_x)).   
\end{equation}
We further assume that the type of a training sample, which could be either collocation, boundary, or initial, is unchanged after perturbing the point.
Note that in the pure AT, one uses the original label for the input ${\mathbf x}$ while training on the adversarial examples. The objective function for PIAT is summarized as:
\begin{equation}\label{PIAT}
\theta^* = \argmin_\theta \ \mathbb{E}_{(({\mathbf x}, t),y)\in\chi} \ \left(\max_{\delta \in [-\epsilon, \epsilon]^{d+1}} \ell({\mathbf x}+\delta_x, t+\delta_t, y'; f_\theta)\right),
\end{equation}
where the loss function is set according to Eq. \ref{loss_func}. PIAT is represented schematically in Fig. \ref{fig:method}.



\section{Numerical examples\label{sec:examples}}
In this section, the adversarial training of physics informed neural network, weight decay and Gaussian smoothing are applied for solving linear and nonlinear equations. Also, the Kuramoto-Sivashinsky equation, Sawada–Kotara equation, and a high dimensional Allen-Cahn equation are considered to conduct more comprehensive analysis.

The training and test points of the evaluation scheme are chosen randomly using Latin Hypercube sampling strategy on the domain of the problem. In order to show the efficiency and capability of the proposed method, the numerical approximations of the mentioned methods are compared. The convergence of our method is also illustrated numerically.

\vspace{7pt}
\noindent
\textbf{Example 1.}

\begin{figure}[b]
    \centering
    \includegraphics[scale=0.6]{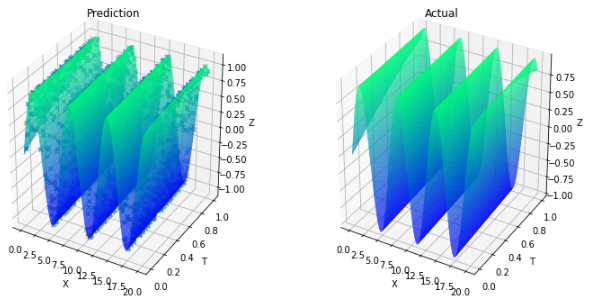}
    \caption{A visual depiction of the actual solution (right), the training points, and predicted solution by PIAT (left) (Example 1).}
    \label{fig:ks_form}
\end{figure}

Let us consider the periodic initial value problem for the fourth-order nonlinear Kuramoto–Sivashinsky (KS) equation as follows:
\begin{equation}\label{eq3}
\begin{split}
&u_t+u u_x+u_{xx}+ \nu u_{xxxx}=f(x, t), \quad (x, t) \in \mathbb{R} \times \mathbb{R}_0^+,\\
&u(x, 0)=g(x),
\end{split}
\end{equation}
where $u=u(x, t)$ is real-valued function and $u^0 : \mathbb{R} \longrightarrow \mathbb{R}$ is sufficiently smooth. 
$u^0$, $f$, and $g$ are $2\pi$-periodic functions and $\nu$ is a positive parameter that is playing the role of viscosity. The solution of Eq. \ref{eq3} is also $2\pi$-periodic in the space domain, i.e. $u(x + 2\pi, t) = u(x, t)$ for all $x \in \mathbb{R}$ and $t \geq 0$. The exact solution of Eq. \ref{eq3} is given by $u(x, t)=\sin(x+t)$.  In all experiments, $T$ is equal to $1$, and $\nu=\frac{1}{2}$. 
A visual depiction of the problem is shown in right panel of Fig. \ref{fig:ks_form}.

By using the proposed methods, the solution is approximated by a deep neural network with 5 hidden layers and 100 neurons per hidden layer and 200 boundary points. The Gaussian smoothing is only applied to the PINN, but the weight decay is applied to both PINN and PIAT. The results are presented in the Table \ref{tab1.4}. First of all, the results demonstrate the effectiveness of PIAT method over PINN regardless of using the weight-decay method. However, the weight-decay improves the results for both PINN and PIAT that demonstrates the advantages of using regularization methods in both cases. Note that the regularization is originally done to prevent overfitting, but in our case this technique has helped to reduce the training error (see the first row of the Table \ref{tab1.4}). We hypothesize that the regularization would make the loss function smoother, and increase the gradient-based optimization effectiveness. Gaussian smoothing, on the other hand, tries to achieve the same goal in an {\it indirect} and {\it blind} way, which does not help with the gradient-based optimization. Adversarial sample augmentation in PIAT, moves the samples {\it purposefully} in the steepest direction that causes large changes in the solution. Therefore, similar to the weight decay, this technique enhances the gradient, in an input adaptive manner, which leads to a better optimization.


\begin{table*}
\setlength{\tabcolsep}{2.8pt}
\caption{Training and test loss functions of standard training with and without weight decay with
$N_u=20$, $N_f=200$, and $\epsilon=0.05$ for $5$ hidden layers, $100$ neurons and $10000$ epochs (Example 1).}
\label{tab1.4}
\begin{center}
\begin{sc}
\begin{tabular*}{\textwidth}{@{\extracolsep{\fill}}*{6}{c}}
\hline\noalign{\smallskip}

 Method &  \multicolumn{2}{c}{PINN}  &  \multicolumn{2}{c}{PIAT} & Gaussian \\

\cline{2-3}\cline{4-5}\cline{6-6}\noalign{\smallskip}
{weight decay} & $\lambda=0$ & $\lambda=5 \times 10^{-4}$ & $\lambda=0$ & $\lambda=5 \times 10^{-4}$ & $\lambda=0$\\
\hline\noalign{\smallskip}

train  	& $1.04 \times 10^{-3}$ & $3.07 \times 10^{-5}$ & $2.23 \times 10^{-6}$ & $1.46 \times 10^{-6}$ & $1.11 \times 10^{-2}$ \\
test 	& $1.08 \times 10^{-3}$ & $2.39 \times 10^{-5}$ & $2.59 \times 10^{-6}$ &  $\mathbf{1.29 \times 10^{-6}}$ & $1.07 \times 10^{-2}$ \\
\hline\noalign{\smallskip}
\end{tabular*} 
\end{sc} 
\end{center}
\end{table*}


\begin{table}
\centering
\caption{Training and test losses, in odd and even rows respectively, for different numbers of boundary points $N_u=20$, $N_f= 200$ and $N_{test}= 100$ for $20000$ epochs (Example 1).}
\label{tab1.1}
\begin{tabular}{ccc}
\hline\noalign{\smallskip}
$N_u$ & PINN & PIAT\\
\noalign{\smallskip}\hline\noalign{\smallskip}
$20$  	& $1.04 \times 10^{-3}$&  	${2.23 \times 10^{-6}}$ \\
  	    & $1.08 \times 10^{-3}$ &  	$\mathbf{2.59 \times 10^{-6}}$ \\
\hline\noalign{\smallskip}
$40$  	& ${5.65 \times 10^{-6}}$ & $4.50 \times 10^{-6}$ \\
  	& $4.87 \times 10^{-6}$ &  	$\mathbf{3.98 \times 10^{-6}}$ \\
\hline\noalign{\smallskip}
$70$   	& $3.88 \times 10^{-5}$ &	$9.25 \times 10^{-6}$ \\
  	& $3.21 \times 10^{-5}$ &  	$\mathbf{8.90 \times 10^{-6}}$ \\
\hline\noalign{\smallskip}
$100$   & $5.90 \times 10^{-6}$& $1.11 \times 10^{-6}$\\
  	& $8.72 \times 10^{-6}$ &  	$\mathbf{9.90 \times 10^{-7}}$ \\
\hline\noalign{\smallskip}
\end{tabular}
\end{table}

\begin{table}
\centering
\caption{Training and test losses, in odd and even rows respectively, for different numbers of neurons and $5$ layers, boundary points $N_u=20$, $N_f= 200$ and $N_{test}= 100$ for $20000$ epochs (Example 1).}
\label{tab1.2}
\begin{tabular}{ccc}
\hline\noalign{\smallskip}
Neuron & PINN & PIAT\\
\noalign{\smallskip}\hline\noalign{\smallskip}
$10$  	& $1.46 \times 10^{-1}$ &  $1.36 \times 10^{-1}$ \\
 	    & $1.44 \times 10^{-1}$ &  	$\mathbf{1.33 \times 10^{-1}}$ \\
\hline\noalign{\smallskip}
$50$  	& $4.40 \times 10^{-5}$ & $1.06 \times 10^{-4}$ \\
  	& $\mathbf{2.81 \times 10^{-5}}$ &  	$9.91 \times 10^{-5}$ \\
\hline\noalign{\smallskip}
$100$   	& $6.44 \times 10^{-5}$ &	$2.23 \times 10^{-6}$ \\
 	& $5.86 \times 10^{-5}$ &  	$\mathbf{2.59 \times 10^{-6}}$ \\
\hline\noalign{\smallskip}
\end{tabular}
\end{table}

\begin{table}
\centering
\caption{Training and test losses, in odd and even rows respectively, for different numbers layers, $50$ Neurons, boundary points $N_u=20$, $N_f= 200$ and $N_{test}= 100$ for $20000$ epochs (Example 1).}
\label{tab1.3}
\begin{tabular}{ccc}
\hline\noalign{\smallskip}
Layer & PINN & PIAT\\
\noalign{\smallskip}\hline\noalign{\smallskip}
$5$  	& $4.40 \times 10^{-5}$ & $1.06 \times 10^{-4}$ \\
 	& $\mathbf{2.81 \times 10^{-5}}$ &  	$9.91 \times 10^{-5}$ \\
\hline\noalign{\smallskip}
$10$  	& $1.50 \times 10^{-3}$ & $3.16 \times 10^{-5}$ \\
  	& $1.40 \times 10^{-3}$  &  	$\mathbf{2.50 \times 10^{-5}}$ \\
\hline\noalign{\smallskip}
$20$   	& $1.43 \times 10^{-4}$ &	$6.51 \times 10^{-6}$ \\
  	& $8.88 \times 10^{-5}$ &  	$\mathbf{2.98 \times 10^{-6}}$ \\
\hline\noalign{\smallskip}
\end{tabular}
\end{table}

\begin{figure}
    \centering
    \vskip 0.2in
    \includegraphics[scale=0.65]{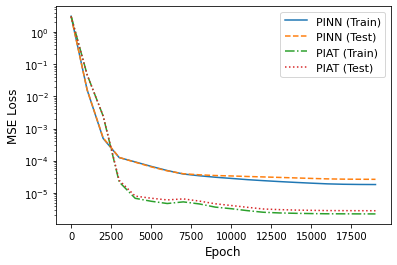}
    \caption{Training and validation losses during the training of the neural network for solving the KS equation using the PINN and PIAT methods (Example 1).}
    \label{fig:ks}
\end{figure}

To check the impacts of the neural network hyper-parameters in the performance of the PIAT method, the training and test errors of the solutions for different numbers of boundary and collocation points have been presented in the Table \ref{tab1.1}. Moreover, Tables \ref{tab1.2} and \ref{tab1.3} show the results for various number of layers and neurons, which all support the general superiority of the PIAT in comparison to the PINN. It is notable that PIAT exhibits this improvement on larger network sizes, which is consistent with the fact that adversarial training requires over-parametrization of the base model \cite{Madry2018}. The train and validation losses during the training is also available for these methods in Fig. \ref{fig:ks}, which demonstrates that both of the methods have converged and trained for sufficient number of epochs.


To check the impacts of the KS problem hyper-parameters in the performance of the PIAT method, the methods are compared within a wider ranges of $x$ and $t$ in Table \ref{tab:KS_wide_range}. Moreover, as another ablation study in this example, the methods are also compared after replacing the exact solution with $u(x,t) = e^{t} . \cos(x) . \sin (1+x)$ and corresponding $f(x, t)$ in Table \ref{tab:KS_new_u}.
All these new findings also point out that PIAT has a much lower error than PINN even with changing the ranges of $x$ and $t$ or the exact solution.

\begin{table}
\centering
\caption{Training and test losses, in odd and even rows respectively, for different ranges of $x$ and $t$ with $N_u=50$, $N_f=2000$, $N_{test}=1000$, 5 layers, 100 neurons, and 20000 epochs (Example 1).}
\label{tab:KS_wide_range}
\begin{tabular}{cccc}
\hline\noalign{\smallskip}
$x$ & $t$ & PINN & PIAT\\
\noalign{\smallskip}\hline\noalign{\smallskip}
$0-20$  	& $0-1$&  	 $4.01\times 10^{-8}$ & 	$2.25\times 10^{-8}$  \\
 	& &  	$6.33 \times 10^{-8}$ & 	$\mathbf{2.36 \times 10^{-8}}$  \\
\hline\noalign{\smallskip}
$0-20$  	& $0-10$&  	 $7.13\times 10^{-7}$ & 	$1.16\times 10^{-7}$ 	  \\
  	& &  	$7.91 \times 10^{-7}$ & 	$\mathbf{1.32 \times 10^{-7}}$  \\
\hline\noalign{\smallskip}
$0-50$  	& $0-1$&  	$9.88\times 10^{-6}$ & 	$5.98\times 10^{-6}$  \\
 	& &  	$2.41 \times 10^{-5}$ & 	$\mathbf{6.29 \times 10^{-6}}$  \\
\hline\noalign{\smallskip}
$0-50$  	& $0-10$&  	 $1.13\times 10^{-4}$ & 	$7.16\times 10^{-5}$ 	  \\
 	& &  	$3.20 \times 10^{-4}$ & 	$\mathbf{7.52 \times 10^{-5}}$  \\
\hline\noalign{\smallskip}
\end{tabular}
\end{table}

\begin{table}
\centering
\caption{Training and test losses for $u(x,t) = e^{t} . \cos(x) . \sin(1+x)$ with $N_u=50$, $N_f=2000$, $N_{test}=1000$, 5 layers,
100 neurons, and 20000 epochs (Example 1).}
\label{tab:KS_new_u}
\begin{tabular}{cccc}
\hline\noalign{\smallskip}
$x$ & $t$ & PINN & PIAT\\
\noalign{\smallskip}\hline\noalign{\smallskip}
$0-20$  	& $0-1$&  	$5.07 \times 10^{-5}$ & 	$4.61 \times 10^{-6}$  \\
 	& &  	$8.30 \times 10^{-5}$ & 	$\mathbf{6.91 \times 10^{-6}}$ \\
\hline\noalign{\smallskip}
\end{tabular}
\end{table}

\vspace{7pt}
\noindent
\textbf{Example 2.}
Sawada-Kotera (SK) equation was widely investigated in many recent works \cite{sk_App_1, sk_App_2}. In this example, we consider to apply PIAT to the seventh order SK equation, which can be shown in the form of:
\begin{equation}\label{eq4}
u_t+(63u^4+63(2u^2 u_{xx}+uu_x^2)+21(u u_{xxxx}+u_{xx}^2+u_x u_{xxx})+u_{xxxxxx})_x=0,
\end{equation}
with the initial condition given by $u(x, 0)=\frac{4k^2}{3} (2-3 \tanh^2(kx))$. The exact solution is:
\begin{equation}\label{eq5}
u(x, t)=\frac{4k^2}{3} (2-3 \tanh^2(k(x-\frac{256k^6}{3}t)).
\end{equation}
This example has also been solved in \cite{13} with the Adomian decomposition method.

The proposed method is applied, and the training and test errors with randomly chosen training and test data have been reported in the Table \ref{tab2.1}, which shows the efficiency of PIAT for solving this kind of problem. Moreover, the results of standard PINN are shown in the Table \ref{tab2.2} and so the efficiency of the PIAT in comparison with the PINN is concluded.

\begin{table*}
\caption{Training and test losses of the PIAT, in odd and even rows respectively, for different numbers of layers and neurons per layer with $N_f=100$, $N_u=10$, $N_{test}=100$ for $10000$ epochs (Example 2).}
\label{tab2.1}
\begin{center}
\begin{sc}
\begin{tabular}{cccc}
\hline\noalign{\smallskip}

 \multirow{2}{*}{Neuron} & \multicolumn{3}{c}{Layers} \\
 \cline{2-4}\noalign{\smallskip}
  &$10$ & $20$ &$40$\\
\noalign{\smallskip}\hline\noalign{\smallskip}
$2$  	& $1.21 \times 10^{-6}$ & ${2.35 \times 10^{-7}}$ 		&  	$3.10 \times 10^{-7}$ \\
     & $1.28 \times 10^{-6}$ & ${2.36 \times 10^{-7}}$ 		&  	$2.74 \times 10^{-7}$ \\
\hline\noalign{\smallskip}
$4$  	& $3.21 \times 10^{-8}$ & $2.13 \times 10^{-8}$      & 	$1.45 \times 10^{-8}$ \\
  	& $2.93 \times 10^{-8}$ & $2.15 \times 10^{-8}$      & 	${1.45 \times 10^{-8}}$ \\
\hline\noalign{\smallskip}
$8$   & $1.24 \times 10^{-8}$ & $ 1.53 \times 10^{-8}$     & 	$1.58 \times 10^{-8}$ \\
  	& $\mathbf{1.27 \times 10^{-8}}$ & $ 1.50 \times 10^{-8}$      & 	$1.57 \times 10^{-8}$ \\
\hline\noalign{\smallskip}
\end{tabular} 
\end{sc} 
\end{center}
\end{table*}

\begin{table*}
	\caption{Training and test losses  of the PINN, in odd and even rows respectively, for different numbers of layers and neurons per layer with $N_f=100$, $N_u=10$, $N_{test}=100$ for $10000$ epochs (Example 2).}
\label{tab2.2}
\begin{center}
\begin{sc}
\begin{tabular}{cccc}
\hline\noalign{\smallskip}

 \multirow{2}{*}{Neuron}  & \multicolumn{3}{c}{Layers} \\
 \cline{2-4}\noalign{\smallskip}
  &$10$ & $20$ &$40$\\
\noalign{\smallskip}\hline\noalign{\smallskip}
$2$  	& $2.73 \times 10^{-6}$ & $7.55 \times 10^{-7}$ 		&  	$7.44 \times 10^{-7}$ \\
     & $2.78 \times 10^{-6}$ & $6.99 \times 10^{-7}$ 		&  	$7.86 \times 10^{-7}$ \\
\hline\noalign{\smallskip}
$4$  	& $4.06 \times 10^{-8}$ & $3.55 \times 10^{-8}$      & 	$1.63 \times 10^{-8}$ \\
	& $3.78 \times 10^{-8}$ & $3.67 \times 10^{-8}$      & 	$1.62 \times 10^{-8}$ \\
\hline\noalign{\smallskip}
$8$   & $1.46 \times 10^{-8}$ & $3.56 \times 10^{-8}$     & 	$2.11 \times 10^{-8}$ \\
  	& $\mathbf{1.46 \times 10^{-8}}$ & $3.46 \times 10^{-8}$      & 	$1.99 \times 10^{-8}$ \\
\hline\noalign{\smallskip}
\end{tabular} 
\end{sc} 
\end{center}
\end{table*}

In Tables \ref{tab2.3} and \ref{tab2.4}, the results of PINN and PIAT with and without weight decay have been presented for different numbers of  neurons. Similar to the example 1, the weight decay improves the results for both PINN and PIAT. Moreover, the combination of PIAT and weight decay reaches the best results in both Tables.

\begin{table*}[t]
\setlength{\tabcolsep}{2.8pt}
\caption{Training and test losses with and without weight decay with
$N_f=100$, $N_u=10$ and $\epsilon=0.05$ for $2$ hidden layers, $10$ neurons and $10000$ epochs training (Example 2).}
\label{tab2.3}
\begin{center}
\begin{sc}
\begin{tabular*}{\textwidth}{@{\extracolsep{\fill}}*{5}{c}}
\hline\noalign{\smallskip}

 Method &  \multicolumn{2}{c}{PINN}  &  \multicolumn{2}{c}{PIAT} \\

\cline{2-3}\cline{4-5}\noalign{\smallskip}
{weight decay} & $\lambda=0$ & $\lambda=5 \times 10^{-4}$ & $\lambda=0$ & $\lambda=5 \times 10^{-4}$\\
\hline\noalign{\smallskip}

train 	& $2.73 \times 10^{-6}$ & $7.64 \times 10^{-8}$ & $1.21 \times 10^{-6}$ & $4.20 \times 10^{-8}$ \\
test  	& $2.78 \times 10^{-6}$ & $6.98 \times 10^{-8}$ & $1.28 \times 10^{-6}$ & $\mathbf{4.06 \times 10^{-8}}$ \\
\hline\noalign{\smallskip}
\end{tabular*} 
\end{sc} 
\end{center}
\end{table*}

\begin{table*}[t]
\setlength{\tabcolsep}{2.8pt}
\caption{training and test loss functions with and without weight decay with
$N_f=100$, $N_u=10$ and $\epsilon=0.05$ for $2$ hidden layers, $20$ neurons  and $10000$ epochs training (Example 2).}
\label{tab2.4}
\begin{center}
\begin{sc}
\begin{tabular*}{\textwidth}{@{\extracolsep{\fill}}*{5}{c}}
\hline\noalign{\smallskip}

 Method &  \multicolumn{2}{c}{PINN}  &  \multicolumn{2}{c}{PIAT} \\

\cline{2-3}\cline{4-5}\noalign{\smallskip}
{weight decay} & $\lambda=0$ & $\lambda=5 \times 10^{-4}$ & $\lambda=0$ & $\lambda=5 \times 10^{-4}$ \\
\hline\noalign{\smallskip}

train  	& $7.55 \times 10^{-7}$ & $1.55 \times 10^{-7}$ & $2.35 \times 10^{-7}$ & $6.78 \times 10^{-9}$ \\
test 	& $6.99 \times 10^{-7}$ & $1.57 \times 10^{-7}$ & $2.36 \times 10^{-7}$ & $\mathbf{6.44 \times 10^{-9}}$ \\
\hline\noalign{\smallskip}
\end{tabular*} 
\end{sc} 
\end{center}
\end{table*}

\begin{figure}
    \centering
    \includegraphics[scale=0.55]{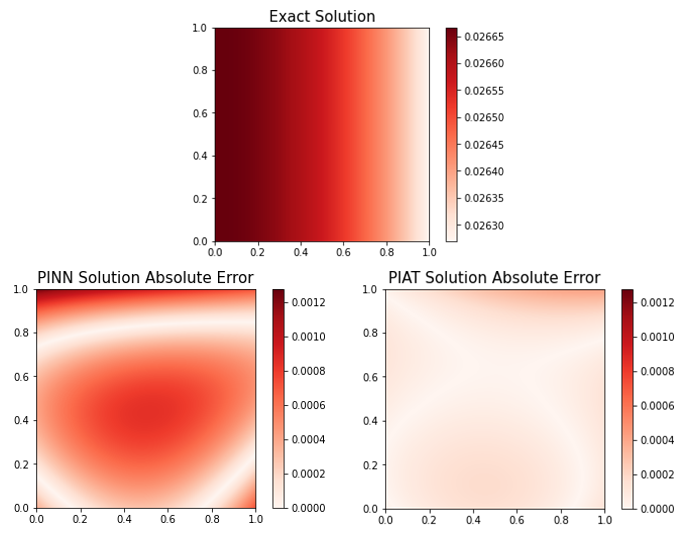}
    \caption{Visualization of exact solution and the point-wise absolute error of predictions by PINN and PIAT ($|f_\theta(x, t) - y|$) in the Example 2. PIAT method achieves much lower error than PINN.}
    \label{fig:errors}
\end{figure}

To perform a visual comparison, a point-wise visualization of absolute errors ($|f_\theta(x, t) - y|$) is presented all over the inputs range in Fig. \ref{fig:errors} for both PINN and PIAT. The results show that PIAT achieves an approximately uniform error all over the inputs range. Moreover, PIAT achieves a much lower errors than PINN, that shows its effectiveness regardless of the choice for test points.

Finally, as an ablation study, the methods are compared within a wider ranges of $x$ and $t$ in Table \ref{tab:SK_wide_range}. These findings also confirms the superiority of PIAT in solving the problem.

\begin{table}
\centering
\caption{Training and test losses, in odd and even rows respectively, for different ranges of $x$ and $t$ with 2 layers,
10 neurons, and 10000 epochs (Example 2).}
\label{tab:SK_wide_range}
\begin{tabular}{cccc}
\hline\noalign{\smallskip}
$x$ & $t$ & PINN & PIAT\\
\noalign{\smallskip}\hline\noalign{\smallskip}
$0-5$  	& $0-5$&  	$6.01 \times 10^{-7}$ & 	$2.32 \times 10^{-7}$  \\
 	& &  	$6.96 \times 10^{-7}$ & 	$\mathbf{3.04 \times 10^{-7}}$  \\
\hline\noalign{\smallskip}
$0-10$  	& $0-10$&  	$4.10 \times 10^{-6}$ & 	$2.83 \times 10^{-7}$  \\
  	& &  	$4.13 \times 10^{-6}$ & 	$\mathbf{3.83 \times 10^{-7}}$  \\
\hline\noalign{\smallskip}
\end{tabular}
\end{table}

\begin{figure}
    \centering
    \includegraphics[scale=0.55]{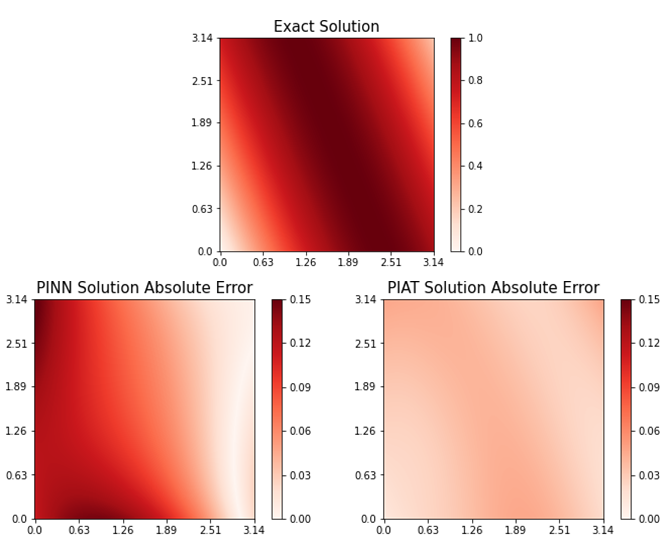}
    \caption{Visualization of exact solution and the point-wise absolute error of predictions by PINN and PIAT ($|f_\theta(x, t) - y|$) in the Example 3 for $d=5$. In order to visualize the solution on a two-dimensional plane, a projection on $x_1$ and $x_2$ dimensions is performed by setting the other dimensions to zero.}
    \label{fig:errors2}
\end{figure}

\vspace{7pt}
\noindent
\textbf{Example 3.}
Consider the high dimensional Allen-Cahn equation as follows:
\begin{equation}\label{eq6}
\begin{split}
u_t& = u_{xx} +u-u^3+ f(x, t), \quad x \in \mathbb{R}^{d} , ~ t \in \mathbb{R},\\
u(0, t)&=0, \quad u(\pi, t)=\sin(\alpha \pi) \cos(2t),\\
u(x,0)& = \sin(\alpha x),\\
\end{split}
\end{equation}
which has the exact solution $u(x, t)=\sin(\alpha x) \cos(2t)$.
The proposed method have been used for solving this problem. The numerical results in Table \ref{tab3.1} show the efficiency of the PIAT in comparison with the standard training of PINN for various dimensions. Moreover, the point-wise visualization of the absolute errors ($|f_\theta(x, t) - y|$) in Fig. \ref{fig:errors2} shows that PIAT has a uniform error all over the domain, which is really important in the real-world problems.

\begin{table*}[t]
\setlength{\tabcolsep}{2.8pt}
\caption{Training and test loss functions of PINN and PIAT with
$N_f=1000$, $N_u=100$, $N_{test}=500$ and $\epsilon=0.05$ (Example 3).}
\label{tab3.1}
\begin{center}
\begin{sc}
\begin{tabular*}{\textwidth}{@{\extracolsep{\fill}}*{7}{c}}
\hline\noalign{\smallskip}
dimension & \multicolumn{2}{c}{$d=1$} & \multicolumn{2}{c}{$d=3$} & \multicolumn{2}{c}{$d=10$} \\
\cline{2-3}\cline{4-5}\cline{6-7}\noalign{\smallskip}

{method} & PINN & PIAT & PINN & PIAT & PINN & PIAT \\
\hline\noalign{\smallskip}
train  	&  $0.01243$ & $0.0060$ & $0.00651$ & $0.00300$ & $0.00269$ & $0.00116$ \\
test  	& $0.01252$ & $\mathbf{0.0059}$ & $0.00641$ & $\mathbf{0.00297}$ & $0.00284$ & $\mathbf{0.00118}$ \\

\hline\noalign{\smallskip}
\end{tabular*} 
\end{sc} 
\end{center}
\end{table*}

Finally, as an ablation study, the methods are compared within a wider ranges of $x$ and $t$ in Table \ref{tab:highdim_wide_range}. The results are similar to previous examples, and show that PIAT has a lower error.

\begin{table}
\centering
\caption{Training and test losses, in odd and even rows respectively, for different ranges of $x$ and $t$ with 5 layers,
80 neurons, 10000 epochs, and $d=2$ (Example 3).}
\label{tab:highdim_wide_range}
\begin{tabular}{cccc}
\hline\noalign{\smallskip}
$x$ & $t$ & PINN & PIAT\\
\noalign{\smallskip}\hline\noalign{\smallskip}
$0-\pi$  	& $0-3$&  	$0.01206$ & 	$0.10180$  \\
 	& &  	$0.01164$ & 	$\mathbf{0.01001}$  \\
\hline\noalign{\smallskip}
$0-2\pi$  	& $0-7$&  	$0.04487
$ & 	$0.00606
$  \\
  	& &  	$0.04654$ & 	$\mathbf{0.00595}$  \\
\hline\noalign{\smallskip}
\end{tabular}
\end{table}

\section*{Conclusion}
In this paper, we have introduced the physics informed adversarial training (PIAT) of neural networks for solving nonlinear problems. Compared to previous works like PINN, our method helps reduce the model’s test errors on various PDE examples, ranging from low-dimensional to high-dimensional problems.  In our simulations, we studied the effects of the different factors on training and test errors, where the convergence of our proposed method was achieved by choosing the different numbers of layers and neurons or input ranges. 

Moreover, we proposed weight decay and Gaussian smoothing which demonstrated the PIAT advantages. The weight decay was applied to both PINN and PIAT, where the obtained results show the efficiency of the PIAT method regardless of the use of the weight decay method compared to the PINN. Also, weight decay improved the results for PINN and PIAT, which demonstrated the benefits of using generalized methods, which PIAT did well. The results of Gaussian smoothing also showed that an adversarial attack works better than a random noise to perturb the samples in PIAT. 

\begin{acknowledgements}
We would like to thank Hossein Yousefi Moghaddam for his insightful comments and helps.
\end{acknowledgements}

{\bf Author's Contributions}: S.S. and M.A. were involved in the conceptualization, formal analysis, investigation, methodology, validation, and drafting the manuscript. M.A. also contributed to the software development and visualization. M.H.R. contributed to the conceptualization, investigation, methodology, supervision, and review/editing of the paper.

{\bf Competing Interests}: The authors have no competing interests to declare. 

{\bf Funding Sources}: This research did not receive any specific grant from funding agencies in the public, commercial, or not-for-profit sectors.
%
%


%
%

\bibliographystyle{ieeetr}
\bibliography{sample.bib}

\begin{thebibliography}{10}

\bibitem{1}
Y.~LeCun, Y.~Bengio, and G.~Hinton, ``Deep learning,'' {\em nature}, vol.~521,
  no.~7553, pp.~436--444, 2015.

\bibitem{2}
Y.~Zhu and N.~Zabaras, ``Bayesian deep convolutional encoderdecoder networks
  for surrogate modeling and uncertainty quantification,'' {\em J. Comput.
  Phys.}, vol.~366, p.~415–447, aug 2018.

\bibitem{3}
Y.~Zhu, N.~Zabaras, P.-S. Koutsourelakis, and P.~Perdikaris,
  ``Physics-constrained deep learning for high-dimensional surrogate modeling
  and uncertainty quantification without labeled data,'' {\em Journal of
  Computational Physics}, vol.~394, pp.~56--81, 2019.

\bibitem{4}
C.~Yang, X.~Yang, and X.~Xiao, ``Data-driven projection method in fluid
  simulation,'' {\em Computer Animation and Virtual Worlds}, vol.~27, no.~3-4,
  pp.~415--424, 2016.

\bibitem{5}
N.~Geneva and N.~Zabaras, ``Quantifying model form uncertainty in
  reynolds-averaged turbulence models with bayesian deep neural networks,''
  {\em Journal of Computational Physics}, vol.~383, pp.~125--147, 2019.

\bibitem{6}
M.~Schöberl, N.~Zabaras, and P.-S. Koutsourelakis, ``Predictive collective
  variable discovery with deep bayesian models,'' {\em The Journal of Chemical
  Physics}, vol.~150, no.~2, p.~024109, 2019.

\bibitem{7}
D.~C. Psichogios and L.~H. Ungar, ``A hybrid neural network-first principles
  approach to process modeling,'' {\em AIChE Journal}, vol.~38, no.~10,
  pp.~1499--1511, 1992.

\bibitem{8}
A.~J. Meade and A.~A. Fernandez, ``The numerical solution of linear ordinary
  differential equations by feedforward neural networks,'' {\em Math. Comput.
  Model.}, vol.~19, p.~1–25, June 1994.

\bibitem{9}
A.~J. Meade and A.~A. Fernandez, ``Solution of nonlinear ordinary differential
  equations by feedforward neural networks,'' {\em Math. Comput. Model.},
  vol.~20, p.~19–44, Nov. 1994.

\bibitem{10}
I.~Lagaris, A.~Likas, and D.~Fotiadis, ``Artificial neural networks for solving
  ordinary and partial differential equations,'' {\em IEEE transactions on
  neural networks}, vol.~9 5, pp.~987--1000, 1998.

\bibitem{11}
M.~W. M.~G. Dissanayake and N.~Phan-Thien, ``Neural-network-based
  approximations for solving partial differential equations,'' {\em
  Communications in Numerical Methods in Engineering}, vol.~10, no.~3,
  pp.~195--201, 1994.

\bibitem{12}
M.~Raissi, P.~Perdikaris, and G.~Karniadakis, ``Physics-informed neural
  networks: A deep learning framework for solving forward and inverse problems
  involving nonlinear partial differential equations,'' {\em Journal of
  Computational Physics}, vol.~378, pp.~686--707, 2019.

\bibitem{14}
D.~Zhang, L.~Lu, L.~Guo, and G.~E. Karniadakis, ``Quantifying total uncertainty
  in physics-informed neural networks for solving forward and inverse
  stochastic problems,'' {\em J. Comput. Phys.}, vol.~397, 2019.

\bibitem{15}
L.~Yang, D.~Zhang, and G.~E. Karniadakis, ``Physics-informed generative
  adversarial networks for stochastic differential equations,'' {\em {SIAM} J.
  Sci. Comput.}, vol.~42, no.~1, pp.~A292--A317, 2020.

\bibitem{16}
X.~Meng and G.~E. Karniadakis, ``A composite neural network that learns from
  multi-fidelity data: Application to function approximation and inverse {PDE}
  problems,'' {\em J. Comput. Phys.}, vol.~401, 2020.

\bibitem{17}
G.~Pang, L.~Lu, and G.~E. Karniadakis, ``fpinns: Fractional physics-informed
  neural networks,'' {\em SIAM J. Sci. Comput.}, vol.~41, pp.~A2603--A2626,
  2019.

\bibitem{18}
Z.~Fang and J.~Zhan, ``A physics-informed neural network framework for pdes on
  3d surfaces: Time independent problems,'' {\em IEEE Access}, vol.~8,
  pp.~26328--26335, 2020.

\bibitem{19}
D.~Zhang, L.~Guo, and G.~E. Karniadakis, ``Learning in modal space: Solving
  time-dependent stochastic pdes using physics-informed neural networks,'' {\em
  SIAM Journal on Scientific Computing}, vol.~42, no.~2, pp.~A639--A665, 2020.

\bibitem{20}
A.~D. Jagtap, E.~Kharazmi, and G.~E. Karniadakis, ``Conservative
  physics-informed neural networks on discrete domains for conservation laws:
  Applications to forward and inverse problems,'' {\em Computer Methods in
  Applied Mechanics and Engineering}, vol.~365, p.~113028, 2020.

\bibitem{Karumuri2020SimulatorfreeSO}
S.~Karumuri, R.~Tripathy, I.~Bilionis, and J.~H. Panchal, ``Simulator-free
  solution of high-dimensional stochastic elliptic partial differential
  equations using deep neural networks,'' {\em J. Comput. Phys.}, vol.~404,
  2020.

\bibitem{21}
E.~Kharazmi, Z.~Zhang, and G.~E. Karniadakis, ``hp-vpinns: Variational
  physics-informed neural networks with domain decomposition,'' {\em Computer
  Methods in Applied Mechanics and Engineering}, vol.~374, p.~113547, 2021.

\bibitem{22}
A.~Jagtap and G.~Karniadakis, ``Extended physics-informed neural networks
  (xpinns): A generalized space-time domain decomposition based deep learning
  framework for nonlinear partial differential equations,'' {\em Communications
  in Computational Physics}, vol.~28, pp.~2002--2041, 11 2020.

\bibitem{23}
L.~Yang, X.~Meng, and G.~E. Karniadakis, ``B-pinns: Bayesian physics-informed
  neural networks for forward and inverse pde problems with noisy data,'' {\em
  Journal of Computational Physics}, vol.~425, p.~109913, 2021.

\bibitem{24}
K.~Shukla, A.~D. Jagtap, and G.~E. Karniadakis, ``Parallel physics-informed
  neural networks via domain decomposition,'' {\em CoRR}, vol.~abs/2104.10013,
  2021.

\bibitem{lu2021deepxde}
L.~Lu, X.~Meng, Z.~Mao, and G.~E. Karniadakis, ``{DeepXDE}: A deep learning
  library for solving differential equations,'' {\em SIAM Review}, vol.~63,
  no.~1, pp.~208--228, 2021.

\bibitem{25}
X.~Meng and G.~E. Karniadakis, ``A composite neural network that learns from
  multi-fidelity data: Application to function approximation and inverse {PDE}
  problems,'' {\em J. Comput. Phys.}, vol.~401, 2020.

\bibitem{26}
A.~D. Jagtap, K.~Kawaguchi, and G.~E. Karniadakis, ``Adaptive activation
  functions accelerate convergence in deep and physics-informed neural
  networks,'' {\em J. Comput. Phys.}, vol.~404, 2020.

\bibitem{27}
S.~Wang, Y.~Teng, and P.~Perdikaris, ``Understanding and mitigating gradient
  flow pathologies in physics-informed neural networks,'' {\em SIAM Journal on
  Scientific Computing}, vol.~43, no.~5, pp.~A3055--A3081, 2021.

\bibitem{28}
L.~Lu, R.~Pestourie, W.~Yao, Z.~Wang, F.~Verdugo, and S.~G. Johnson,
  ``Physics-informed neural networks with hard constraints for inverse
  design,'' {\em ArXiv}, vol.~abs/2102.04626, 2021.

\bibitem{29}
H.~Gao, L.~Sun, and J.-X. Wang, ``Phygeonet: Physics-informed geometry-adaptive
  convolutional neural networks for solving parameterized steady-state pdes on
  irregular domain,'' {\em Journal of Computational Physics}, vol.~428,
  p.~110079, 2021.

\bibitem{30}
Y.~Shin, J.~Darbon, and G.~Em~Karniadakis, ``On the convergence of physics
  informed neural networks for linear second-order elliptic and parabolic type
  pdes,'' {\em Communications in Computational Physics}, vol.~28, no.~5,
  pp.~2042--2074, 2020.

\bibitem{31}
S.~Mishra and R.~Molinaro, ``Estimates on the generalization error of physics
  informed neural networks (pinns) for approximating pdes,'' {\em ArXiv},
  vol.~abs/2007.01138, 2020.

\bibitem{32}
S.~Wang, X.~Yu, and P.~Perdikaris, ``When and why pinns fail to train: {A}
  neural tangent kernel perspective,'' {\em CoRR}, vol.~abs/2007.14527, 2020.

\bibitem{Krogh91}
A.~Krogh and J.~A. Hertz, ``A simple weight decay can improve generalization,''
  in {\em Advances in Neural Information Processing Systems 4, {NIPS}}, 1991.

\bibitem{SzegedyZSBEGF13}
C.~Szegedy, W.~Zaremba, I.~Sutskever, J.~Bruna, D.~Erhan, I.~J. Goodfellow, and
  R.~Fergus, ``Intriguing properties of neural networks,'' in {\em 2nd
  International Conference on Learning Representations, {ICLR}}, 2014.

\bibitem{NguyenYC15}
A.~M. Nguyen, J.~Yosinski, and J.~Clune, ``Deep neural networks are easily
  fooled: High confidence predictions for unrecognizable images,'' in {\em
  {IEEE} Conference on Computer Vision and Pattern Recognition, {CVPR}}, 2015.

\bibitem{Madry2018}
A.~Madry, A.~Makelov, L.~Schmidt, D.~Tsipras, and A.~Vladu, ``Towards deep
  learning models resistant to adversarial attacks,'' in {\em 6th International
  Conference on Learning Representations, {ICLR}}, 2018.

\bibitem{ShortenK19}
C.~Shorten and T.~M. Khoshgoftaar, ``A survey on image data augmentation for
  deep learning,'' {\em J. Big Data}, vol.~6, p.~60, 2019.

\bibitem{Luke18}
L.~Taylor and G.~Nitschke, ``Improving deep learning with generic data
  augmentation,'' in {\em 2018 IEEE Symposium Series on Computational
  Intelligence (SSCI)}, pp.~1542--1547, 2018.

\bibitem{IlyasSTETM19}
A.~Ilyas, S.~Santurkar, D.~Tsipras, L.~Engstrom, B.~Tran, and A.~Madry,
  ``Adversarial examples are not bugs, they are features,'' in {\em Advances in
  Neural Information Processing Systems 32: Annual Conference on Neural
  Information Processing Systems 2019, NeurIPS} (H.~M. Wallach, H.~Larochelle,
  A.~Beygelzimer, F.~d'Alch{\'{e}}{-}Buc, E.~B. Fox, and R.~Garnett, eds.),
  2019.

\bibitem{d2020underspecification}
A.~D'Amour, K.~Heller, D.~Moldovan, B.~Adlam, B.~Alipanahi, A.~Beutel, C.~Chen,
  J.~Deaton, J.~Eisenstein, M.~D. Hoffman, {\em et~al.}, ``Underspecification
  presents challenges for credibility in modern machine learning,'' {\em arXiv
  preprint arXiv:2011.03395}, 2020.

\bibitem{ZhangYJXGJ19}
H.~Zhang, Y.~Yu, J.~Jiao, E.~P. Xing, L.~E. Ghaoui, and M.~I. Jordan,
  ``Theoretically principled trade-off between robustness and accuracy,'' in
  {\em Proceedings of the 36th International Conference on Machine Learning,
  {ICML}}, 2019.

\bibitem{ZhangXH0CSK20}
J.~Zhang, X.~Xu, B.~Han, G.~Niu, L.~Cui, M.~Sugiyama, and M.~S. Kankanhalli,
  ``Attacks which do not kill training make adversarial learning stronger,'' in
  {\em Proceedings of the 37th International Conference on Machine Learning,
  {ICML}}, 2020.

\bibitem{az2021}
M.~Azizmalayeri and M.~H. Rohban, ``Lagrangian objective function leads to
  improved unforeseen attack generalization in adversarial training,'' {\em
  CoRR}, vol.~abs/2103.15385, 2021.

\bibitem{sk_App_1}
J.~Manafian and M.~Lakestani, ``Lump-type solutions and interaction phenomenon
  to the bidirectional sawada--kotera equation,'' {\em Pramana}, vol.~92,
  no.~3, pp.~1--13, 2019.

\bibitem{sk_App_2}
M.~Osman, ``One-soliton shaping and inelastic collision between double solitons
  in the fifth-order variable-coefficient sawada--kotera equation,'' {\em
  Nonlinear Dynamics}, vol.~96, no.~2, pp.~1491--1496, 2019.

\bibitem{13}
D.~Kaya and M.~Aassila, ``An application for a generalized kdv equation by the
  decomposition method,'' {\em Physics Letters A}, vol.~299, no.~2,
  pp.~201--206, 2002.

\end{thebibliography}

\end{document}